\documentclass[sigconf,screen]{acmart}

\usepackage{multirow}
\usepackage{graphicx}
\usepackage{caption}
\usepackage{subcaption}
\usepackage{dblfloatfix}
\usepackage{xcolor}
\usepackage{booktabs}
\usepackage{wrapfig}
\usepackage{enumitem}
\usepackage{cleveref}
\usepackage{longtable}
\usepackage{ragged2e}
\usepackage{array}
\usepackage{amsmath}
\usepackage{pgfplots}
\usepackage{tikz}
\usepackage{siunitx}
\usepackage{color,soul}
\setlist{noitemsep, leftmargin=*, topsep=0pt, partopsep=0pt}
\crefformat{section}{\S#2#1#3} 
\crefformat{subsection}{\S#2#1#3}
\crefformat{subsubsection}{\S#2#1#3}

\usepackage{underscore} 
\catcode`\_=12 
\setcopyright{none}
\acmConference[\ifnum\value{page}=1 HARMONY: Human-centered AI Research for Mental Health, an Open Networking Symposium\else HARMONY 2026\fi]{HARMONY: Human-centered AI Research for Mental Health, an Open Networking Symposium}{August 6, 2026}{Pittsburgh, PA, USA}
\acmYear{2026}
\copyrightyear{2026}

\newcommand*\circled[1]{\tikz[baseline=(char.base)]{\small{\textbf{
			\node[shape=circle,fill,inner sep=0.75pt] (char) {\textcolor{white}{#1}};}}}}

\begin{document}
\newcolumntype{L}[1]{>{\RaggedRight\arraybackslash}p{#1}}
\title{Toward Contemplative LLM: A Modular Framework for Evaluating and Enhancing LLM Alignment in Mental Health}

\author{Asher Sprigler}
\affiliation{%
  \department{Electrical and Computer Engineering}
  \institution{Purdue University}
  \city{West Lafayette}
  \state{IN}
  \country{USA}
}
\email{asprigle@purdue.edu}

\author{Yang-Yang Feng}
\affiliation{%
  \department{Biomedical Engineering}
  \institution{Washington University in St. Louis}
  \city{St. Louis}
  \state{MO}
  \country{USA}
}
\email{yang-yang.feng@wustl.edu}

\author{Iftach Amir}
\affiliation{%
  \department{Psychological \& Brain Sciences}
  \institution{Washington University in St. Louis}
  \city{St. Louis}
  \state{MO}
  \country{USA}
}
\email{amiri@wustl.edu}

\author{Jonathan E. Bogard}
\affiliation{%
  \department{Olin Business School}
  \institution{Washington University in St. Louis}
  \city{St. Louis}
  \state{MO}
  \country{USA}
}
\email{bogard@wustl.edu}

\author{Todd S Braver}
\affiliation{%
  \department{Psychological \& Brain Sciences}
  \institution{Washington University in St. Louis}
  \city{St. Louis}
  \state{MO}
  \country{USA}
}
\email{tbraver@wustl.edu}

\author{Yi Ding}
\affiliation{%
  \department{Electrical and Computer Engineering}
  \institution{Purdue University}
  \city{West Lafayette}
  \state{IN}
  \country{USA}
}
\email{yiding@purdue.edu}

\author{David Kinney}
\affiliation{%
  \department{Department of Philosophy}
  \institution{Washington University in St Louis}
  \city{St. Louis}
  \state{MO}
  \country{USA}
}
\email{kinney@wustl.edu}

\author{Yixue Zhao}
\affiliation{%
  \department{Interdisciplinary Science}
  \institution{Yixue Research Institute}
  \city{Washington}
  \state{DC}
  \country{USA}
}
\email{yixue@yixuezhao.com}

\maketitle


\section{Introduction}\label{sec:intro}
Large language models (LLMs) are rapidly transforming human-AI interaction, enabling new capabilities in communication, decision-making, and knowledge generation. However, these advances introduce critical challenges in ethical alignment, as models evolve rapidly toward increasingly autonomous behavior that may surpass human-level intelligence~\cite{ji2023ai,naveed2025comprehensive}. This trajectory underscores the urgency of proactively guiding AI development to ensure alignment.

Contemplative traditions have long guided ethical behavior and prosocial interaction. In modern contexts, these traditions have been adapted into scientific paradigms for studying cognitive training, therapeutic
intervention, and neuroscience investigations~\cite{berryman2023contemplative,ozawa2016contemplative}. Research on contemplative practices such as mindfulness and meditation shows effects on emotional regulation, cognition, and moral reasoning~\cite{kirk2016mindfulness,pless2017mindfulness,shapiro2012mindfulness,urrila2022personal}. Recent work further suggests that contemplative principles (e.g., mindfulness, compassion, non-dual reasoning), grounded in both ancient traditions and modern science, may offer a promising alternative paradigm for alignment. Early evidence indicates that incorporating such principles into prompting strategies can improve cooperation and reduce ethical violations in LLM outputs~\cite{laukkonen2025contemplative}, suggesting the potential for a contemplatively based constitutional approach to ethical alignment.

However, as new models, evaluation metrics, and benchmarks emerge rapidly, it remains challenging to systematically assess whether and how contemplative principles enhance LLM alignment across diverse and evolving scenarios. In many cases, newly released models can render prior experimental findings obsolete almost immediately. Existing approaches are often developed as \textit{ad hoc}  solutions, which lack the flexibility to adapt quickly to new models, metrics, or benchmarks. Clear examples can be seen in current state-of-the-art work in the mental health domain where LLM alignment is crucial~\cite{badawi-etal-2026-trust,xu2025mentalchat16k,li2026counselbench}. Thus, it is vital to develop generalizable approaches to quickly test new models against a wide range of potential benchmarks and metrics.

To address this challenge and establish a strong foundation for future work on contemplative LLM, we have designed a modular framework that enables seamless integration of new models, metrics, and benchmarks through a reusable and customizable pipeline, initially targeted towards the mental health domain. Currently, our framework can reproduce existing state-of-the-art results~\cite{xu2025mentalchat16k,badawi-etal-2026-trust,li2026counselbench} and support systematic cross-evaluation by flexibly ``mixing and matching'' models, metrics, and benchmarks, enabling fair comparison and deeper insights. Building on these preliminary results, our future work will leverage the framework to investigate how contemplative principles enhance LLM alignment across other domains, including decision-making, moral reasoning, and therapeutic interactions. The central goal is to bridge computational evaluation with human-centered ethical reasoning by integrating interdisciplinary aspects such as cognitive science, behavioral economics, philosophy, and system design.

\section{Framework Core Features}\label{sec:framework}
In the mental health domain, LLM evaluations are often ad hoc and lack standardization, hindering systematic comparisons across models and benchmarks. Existing pipelines are also tightly coupled to specific tasks or metrics, limiting extensibility as new models and criteria emerge. Moreover, prior work rarely incorporates prompting-based interventions (e.g., contemplative principles~\cite{laukkonen2025contemplative}) as a systematic approach to improving model behavior~\cite{badawi-etal-2026-trust,li2026counselbench,pombal2025mindeval,xu2025mentalchat16k}.

To address these limitations, we design a modular and extensible evaluation framework with three core features.
\circled{1} \textbf{Flexible and Modular Pipeline (Implemented):} The framework supports reusable and customizable metrics, models, and benchmarks, largely reducing development effort. For instance, researchers can define alignment metrics (e.g., ethical consistency, empathy, cooperation) and apply them uniformly across models and tasks, enabling standardized and rapid evaluation.
\circled{2} \textbf{Cross Evaluation Support (Implemented):}
By decoupling the evaluation mechanism from specific tasks, the framework enables systematic comparison across multiple LLMs and benchmarks, such as applying metrics from one benchmark to another. This provides insights into how models perform under different alignment criteria and which are best suited for specific downstream applications.
\circled{3} \textbf{Plug-and-Play Prompting Module (Under Development):} Prompting techniques (e.g., contemplative principles~\cite{laukkonen2025contemplative}) are designed as a separate module that can be easily modified or replaced without altering the core pipeline, enabling rapid comparison across models and tasks. This design supports interdisciplinary collaboration by allowing domain experts to define principles that can be automatically translated into structured prompts and evaluated within the pipeline, without requiring technical expertise. This ``plug-and-play'' design allows users to focus on their domain expertise while leveraging a shared computational infrastructure.

\section{Our Vision}\label{sec:vision}
The proposed framework advances LLM alignment by providing a standardized, modular, and extensible infrastructure for evaluating and enhancing LLM behavior. By enabling rapid and systematic evaluation across models, metrics, and benchmarks, it addresses a key bottleneck in the current landscape where fragmented, ad hoc approaches have impeded reproducibility and progress. Its plug-and-play prompting modules provide a principled pathway for incorporating ethical perspectives (e.g., contemplative principles~\cite{laukkonen2025contemplative}), fostering interdisciplinary development of human-aligned AI systems.
Although initially focused on mental health, the framework is domain-agnostic and can extend to areas such as decision-making, moral reasoning, and human-AI collaboration. This work integrates human behavioral insights into AI systems, supporting the development of robust, trustworthy, and socially beneficial human-AI ecosystems. Future work will use the framework to explore diverse ethical approaches, including contemplative principles, to improve LLM behavior beyond mental health.


\bibliographystyle{ACM-Reference-Format}
\bibliography{ref}

\end{document}